%% file: main.tex
\documentclass[letterpaper]{article} 
\usepackage{aaai25}  
\usepackage{times}  
\usepackage{helvet}  
\usepackage{courier}  
\usepackage[hyphens]{url}  
\usepackage{graphicx} 
\usepackage{natbib}  
\usepackage{caption} 
\usepackage{algorithm}
\usepackage{algorithmic}
\usepackage{amssymb}
\usepackage{amsmath}
\usepackage{amsthm}
\newtheorem{theorem}{Theorem}[section]

\usepackage{booktabs}
\usepackage{float}

\usepackage{newfloat}
\usepackage{listings}
\DeclareCaptionStyle{ruled}{labelfont=normalfont,labelsep=colon,strut=off} 
\floatstyle{ruled}
\newfloat{listing}{tb}{lst}{}
\floatname{listing}{Listing}
\title{DILEMMA: Joint LLM Quantization and Distributed LLM Inference Over Edge Computing Systems}
\author{
    Minoo Hosseinzadeh, \textsuperscript{\rm 1}
    Hana Khamfroush, \textsuperscript{\rm 1}
}
\affiliations{
 \textsuperscript{\rm 1} University of Kentucky\\
    mho357@uky.com, khamfroush@uky.com
}

\usepackage{bibentry}

\begin{document}

\maketitle

\begin{abstract}
With a recent trend of using Large Language Models~(LLMs) for different applications within smart cities, there is a need for pushing these models toward the edge of network while still preserving their performance. Edge Computing~(EC) as a physically closer computing resource to the end users can help to reduce the communication delay for serving end users' tasks for LLM-dependent services. However, EC servers have limited capacity in terms of communication, computation, and storage capacity.
This paper introduces DILEMMA, a novel framework addressing the challenges of deploying LLMs in EC systems by jointly optimizing layer placement and layer quantization in EC systems. DILEMMA formulates an Integer Linear Programming problem to minimize total inference delay while ensuring acceptable LLM performance levels, leveraging layer-wise quantization and knowledge distillation for LLM performance control. Experimental evaluations on OPT-350 model using the SQuAD dataset demonstrate that DILEMMA achieves a quantization ratio of up to 12.75\% while preserving model loss, highlighting its effectiveness in resource-constrained environments.
\end{abstract}

\section{Introduction}
\label{sec:introduction}
\input{Sections/introduction}


\section{Problem Definition}
\label{sec:Problem Definition}
\input{Sections/problem_definition}

\section{Problem Formulation}
\label{sec:Problem Formulation}
\input{Sections/problem_formulation}

\section{Results}
\label{sec:Results}
\input{Sections/results}

\section{Conclusion}
\label{sec:Conlusion}
\input{Sections/conclusion}

\section*{Acknowledgment}
\noindent This work is funded by research grants provided by the National Science Foundation~(NSF) under the grant number 2340075.

\input{main.bbl}

\end{document}

%% file: Sections/Introduction.tex
Smart City applications~\cite{gaur2015smart} heavily rely on Machine Learning~(ML) models~\cite{hosseinzadeh2020optimal, hosseinzadeh2021joint} for their various services. Typically, these applications leverage Edge Computing~(EC)~\cite{ETSI} as their underlying computational infrastructure~\cite{hosseinzadeh2020optimal, hosseinzadeh2021joint}.
The recent progress in Large Language Models~(LLM)~\cite{vaswani2017attention, radford2018improving} impacts numerous applications within smart cities~\cite{krystek2024managing}. Notably, several applications have seen significant improvements in performance, particularly in areas such as and smart virtual assistants~\cite{sezgin2024redefining}.
However, the computational demands of LLM models pose challenges, particularly when it comes to running them on end-user devices or even EC devices~\cite{bhardwaj2024survey}.

EC devices have limited communication, computation, and storage capacity. Several smart city applications are computationally intensive to run on limited capacity devices such as EC devices~\cite{hosseinzadeh2020optimal, hosseinzadeh2021joint}. Research has shown that EC systems provide a great opportunity for distributed service running in the smart cities~\cite{karjee2022split}. This option can be used for collaborative inference systems for ML-dependent services~\cite{li2024distributed} and more specifically LLM-dependent services~\cite{cai2024edge, zhao2024llm}.

On the other hand, ML model quantization techniques~\cite{gholami2022survey} have been widely used to reduce the ML models size. Additionally, this size reduction helps to reduce the memory cost and computational cost associated with the inference process, making it feasible to run these models across a variety of devices with varying computational capacities~\cite{gholami2022survey}. Model quantization transforms the model parameters from float32 or float64 to lower-bit float numbers or even integer numbers. The fewer the bits, the less memory is used. It also helps with fastening the inference time~\cite{cai2024edge}. LLM quantization can be done layer-wise, channel-wise, row/column-wise, token-wise, and group-wise~\cite{xiao2023smoothquant}. 

Research shows the benefit of either quantization~\cite{xiao2023smoothquant, bai2022towards, li2023fptq} or distributed inference~\cite{wu2023fast, borzunov2024distributed, zhao2024lingualinked, zhang2024edgeshard} to improve LLM inference time. There are small number of papers that investigated the joint quantization and distributed LLM inference problem~\cite{cai2024edge, zhao2024llm}. Cai et al.~\cite{cai2024edge} proposed a client-server-based LLM inference using EC servers while quantizing the LLM. The difference between our paper and \cite{cai2024edge} is that we consider a fully distributed system model.
Zhao et al.~\cite{zhao2024llm} proposed a joint distributed inference and LLM quantization method to reduce the total delay. The authors distribute the LLM layers over a set of GPU servers while quantizing each layer of the LLM. The main difference between our work and ~\cite{zhao2024llm} is first the objective functions and the system models are different. Second, the authors in paper~\cite{zhao2024llm} uses the variance of output of a layer due to weight quantization to estimate the performance of the LLM after quantization while we use a different approach.

In this paper, we consider a two-tier EC system consisting of a heterogeneous set of edge servers and a cloud server where EC servers are connected to each other using Device-to-Device~(D2D) communication as represented in Fig.~\ref{fig:system_arch}. EC servers have limited communication, computation, and storage capacity. We assume that there are LLM-dependent tasks offloaded to the EC servers. And EC servers do not have enough capacity to run the entire LLM by themselves. Additionally, tasks are time-sensitive. Therefore, it is preferred to run the tasks on the edge devices compared to offloading them to the cloud server. 
We propose a \textbf{DI}stributed \textbf{L}LM plac\textbf{EM}ent and layer-wised LL\textbf{M} qu\textbf{A}ntization scheme~(DILEMMA) in EC systems while simultaneously making decision on layer-wise quantization of LLM models to reduce the total cost of the system. To solve this problem, we propose an optimization model to jointly decide which edge server host adn run which layer of the LLM while making decision on each layer quantization independent of other layers. We summarize the contributions of this work as following:

\begin{itemize}
    \item We investigate the problem of joint layer placement on the edge computing servers and layer quantization to minimize the total completion time--called DILEMMA.
    \item We formulate the DILEMMA problem as an Integer Linear Programming~(ILP) model and prove that the DILEMMA model is NP-hard when the number of LLMs in the system is greater than one.
    \item We solve the DILEMMA model for the case when we only have one LLM model in the system using off-the-shelf optimization solver and show that using our proposed method to control the LLM performance we can compress the model by $12.75\%$ while still preserving a good level of LLM performance.
\end{itemize}

%% file: Sections/Problem_definition.tex
\begin{figure}
    \centering
    \includegraphics[width=0.9\linewidth]{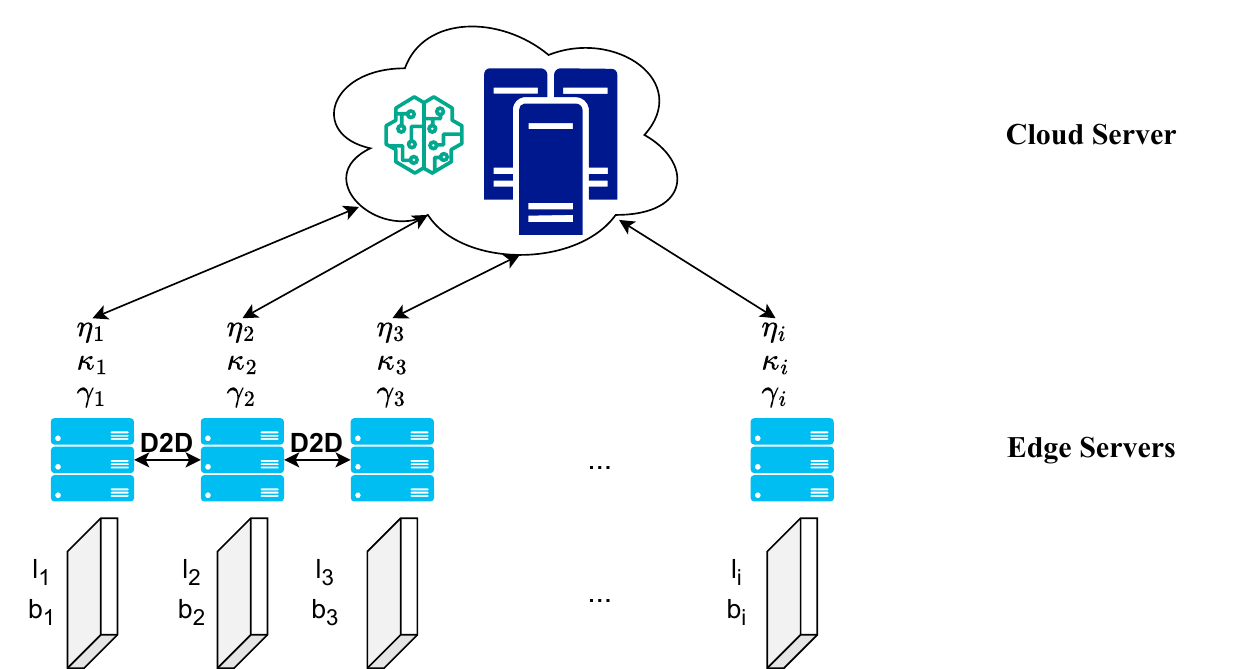}
    \caption{Two Layers System Architecture}
    \label{fig:system_arch}
\end{figure}

\textbf{System Definition.} 
We consider a two layers system as represented in Fig.~\ref{fig:system_arch}. The first layer consists of the cloud server $cl$. The cloud server has the full knowledge of the system and the DILEMMA decision making happens on the cloud server.
The second layer is the edge layer consisting of a set of EC servers, $\mathcal{M}$. Edge servers are heterogeneous in terms of hardware capacity (e.g. communication, computation, and storage capacity). Each edge server $i \in \mathcal{M}$ has limited communication capacity, $\eta_i$, computation capacity, $\gamma_i$, and storage capacity, $\kappa_i$. EC servers are connected to their neighborhood edge servers using Device-to-Device~(D2D) communication from one side and they are connected to the cloud server from another side. 

The LLM consists of a set of layers represented by the set of $\mathcal{L}$. For each layer $l \in \mathcal{L}$, it is possible to quantize its weights with $b$ bit size. We show the possible bit sizes with the set $\mathcal{B}$. To limit the range of set $\mathcal{B}$, we define two input variables, $b_{min}$ and $b_{max}$, representing the minimum number of bits that each layer weights can have and the maximum number of bits that each layer weights can have, respectively. Therefore, $\mathcal{B} = \{b_{min}, b_{min}+1, b_{min}+2, ..., b_{max}\}$. Please note that the $b_{max}$ is equal to the number of bits of the weights that the LLM is trained with. 
The goal of DILEMMA problem is to make two important decisions: 1) each layer of LLM should be placed on which edge server, and 2) how much should we quantize each layer which depends on the capacity of that edge server and the desired performance of LLM.

%% file: Sections/Problem_formulation.tex
The objective of the DILEMMA problem is to minimize the total delay incurred by inference time of each layer placed on a specific edge server, and sending the output of each layer from one edge server to the edge server hosting the next layer while still preserving a satisfying level of LLM performance.
We define two decision variables to jointly decide both the placement of the LLM layers on edge servers and the degree of quantization for each placed layer. The first decision variable is $x_{ilb} \triangleq 1$ when the layer $l \in \mathcal{L}$ is placed on the edge server $i \in \mathcal{M}$ with layers quantized with $b$ bits. The second decision variable is $y_{ij}^s$ which is a binary variable representing the output of edge server $i$ will be sent to the edge server $j$ assuming that $x_{ilb} = 1$ and $x_{j(l+1)b} = 1$. 

When the parameters of each layer $l \in \mathcal{L}$ are quantized using $l$ bits, the edge server $i \in \mathcal{M}$ hosting layer $l$ needs a minimum computational capacity to process this layer. We show this parameter using $cp_{ilp}$ which is dependent on the number of operations in layer $l$, the number of bits $b$ for each weight~(i.e. precision), and the FLOPS of edge server $i$. We explain how to calculate $cp_{ilp}$ in Sec.~\ref{section:EstimatingCompletionTime}.
Now, we define the DILEMMA optimization problem as following. 

\begin{equation}
    T = \sum_{i \in \mathcal{M}} \sum_{l \in \mathcal{L}} {x_{ilb}} \; cp_{ilb} + y_{ij} \; cm_{ilj}
\end{equation}

\noindent where the objective is to minimize the total completion time $T$ considering the following constraints. 

\begin{subequations}
\small
    \label{eq:optimization}
    \begin{align}
        \min_{x_{ilb}, y_{ij}}~ 
        & {T}
        \label{eq:objective}
        \\
        \text{s.t.: } \\
        & \sum_{i \in \mathcal{M}} \sum_{l \in \mathcal{L}} {x_{ilb} = 1} \; \qquad \quad \forall l \in \mathcal{L}, \; \forall b \in \mathcal{B}
        \label{eq:constraintplacement}
        \\
        & \sum_{i \in \mathcal{M}} \sum_{l \in \mathcal{L}} {x_{ilb} \leq 1} \; \qquad \quad \forall i \in \mathcal{M} 
        \label{eq:constraintserving}
        \\
        & str_{ilb} \; x_{ilb} \; p_l \leq \kappa_i 
        \qquad \quad \forall i\in \mathcal{M}, l \in \mathcal{L}, \forall b\in \mathcal{B}
        \label{eq:constraintstorage}
        \\
        & x_{ilb} + x_{j(l+1)b} - 1 \leq y_{ij}  \quad \forall i \& j \in \mathcal{M}, l \in \mathcal{L}, \forall b\in \mathcal{B}
        \label{eq:dependencyVariable}
        \\
        & |P - P_{lb}| \leq \epsilon, 
       \qquad \forall i\in \mathcal{M}, l\in \mathcal{L}, \forall b\in \mathcal{B}
        \label{eq:PerformanceGuarantee}
        \\
        & x_{ilb} \& y_{ij} \in {\{ 0,1\}}, 
       \qquad \forall i\in \mathcal{M}, l\in \mathcal{L}, \forall b\in \mathcal{B}
        \label{eq:XdecVariables}
    \end{align}
\end{subequations}

\noindent where constraint~\eqref{eq:constraintplacement} guarantees that each layer $l$ of LLM is placed only at one edge server $i$ and is quantized with $b$ bits. Constraint~\eqref{eq:constraintserving} guarantees that each EC server only hosts one layer $l \in \mathcal{L}$ or does not host any layer.
Eq.~\eqref{eq:constraintstorage} guarantees that edge server $i$ has enough storage capacity to host the layer $l$ quantized with $b$ bits. The constraint~\eqref{eq:dependencyVariable} ensures that if layer $l$ is placed on edge server $i$ and the layer $l+1$ is placed at edge server $j$, then the decision variable $y_{ij}$ is equal to one~(i.e. to consider the dependency between layers). The constraint~\eqref{eq:PerformanceGuarantee} guarantees that the difference between performance of LLM model after quantizing all layers and performance of LLM model before quantization is less than an error rate, $\epsilon$. The last constraint emphasizes on that both $x_{ilb}$ and $y_{ij}$ are binary decision variables.

\textbf{Estimating Completion Time.}
\label{section:EstimatingCompletionTime}
We define the total completion time considering the processing delay and communication delay for each layer. Note that most LLMs use autoregressive inference. It means that the model processes one token at a time, passes it to the entire model, and then generates the next token and repeats the process. Therefore, with a number of tokens of $n$, there will be $n$ communication rounds and computation rounds. Additionally, we assume that every edge server stores its past attention cache, so that in every round each server only transfers activation for a single token.

\textbf{Estimating $\textit{cp}_{ilp}$.}
To estimate the computation delay, $\textit{cp}_{ilp}$, of running the inference task on edge server $i$, we use the FLoating-point Operations per Second~(FLOPS). We model the $\textit{cp}_{ilp}$ as following:

\begin{equation}
    \textit{cp}_{ilp} = n \times \frac{\textit{FLOPS}_l}{\textit{CCS}_i} \times \frac{b \; p_l}{\textit{Original Bit Precision}_l}
\end{equation}

\noindent where $\textit{FLOPS}_l$ represents the number of floating-point operations required for the inference task of layer $l$. We assume FLOPS as a function based on the type of the layer, the number of input features, the number of output features, kernel size~(if applicable), and the hidden state size~(if applicable). 
$\textit{CCS}_i$ is the CPU clock speed of edge server $i$. $b$ is the number of bits of layer $l$ after quantization. $p_l$ is the total number of output tensor of layer $l$. And, $\textit{Original Bit Precision}_l$ represents the bit precision of the original parameters of layer $l$~(i.e. before quantization). $n$ is the number of tokens. We multiply it by $n$ to consider the autoregressive inference behavior which requires $n$ pass through the LLM.

\textbf{Estimating $cm_{ilb}$.}
We estimate the communication delay due to send the output of layer $l$ placed on the edge server $i$ represented with $b$ bits to the next server considering the upload delay. We ignore the download delay on the next server because it usually takes less amount of communication capacity to download compared to upload~\cite{hosseinzadeh2020optimal}. We consider the transmission delay and propagation delay to model the communication delay to transmit data from edge server $i$ to the edge server $j$ as:

\begin{equation}
    \textit{cm}_{ilp} = n \times \frac{p_l \; b_l}{\Omega_{ij}}
\end{equation}

\noindent where $\Omega_{ij}$ is the link capacity between EC server $i \in \mathcal{M}$ and EC server $j \in \mathcal{M}$. $p_l$ is the total number of output tensor of layer $l$ and $b_l$ is the precision~(number of bits) of layer $l$. The total number of output tensor of layer $l$ depends on the number of tokens $n$, batch size, and the embedding size. We represent the batch size by $\beta$, and embedding size by $e$. And, $n$ is the number of tokens.

\begin{figure}
    \centering
    \includegraphics[width=0.9\linewidth]{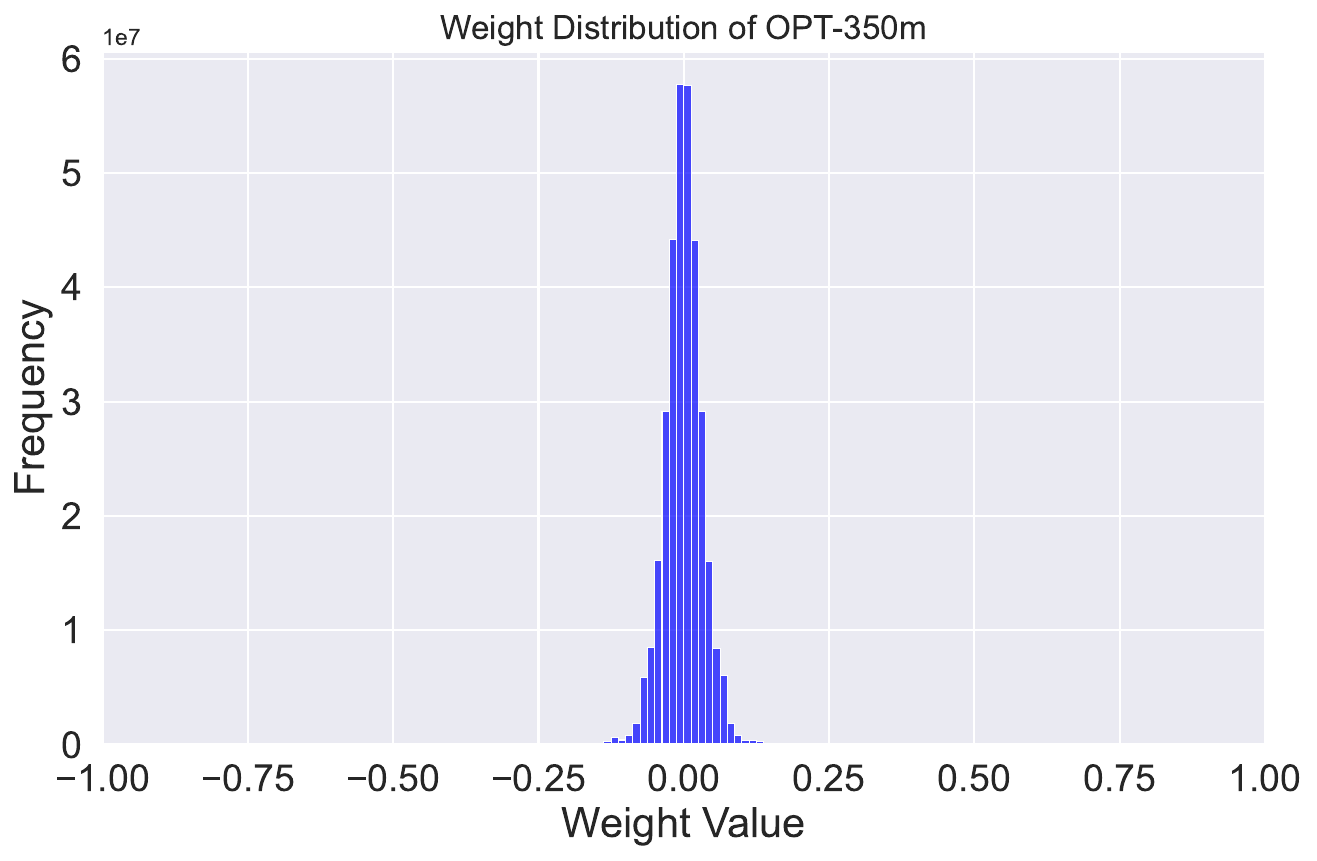}
    \caption{Distribution of the OPT-350m Weights}
    \label{fig:opt350_weight_dist}
\end{figure}

\textbf{Estimating LLM Performance After Layer-wise Quantization.}
The main challenge of solving the DILEMMA problem is the constraint~\eqref{eq:PerformanceGuarantee}. The authors of recent paper~\cite{lin2024awq} have used loss value as the evaluation metric of the LLM performance after quantizing each layer. However, the value of loss is not the only metric to evaluate the LLMs~\cite{chang2024survey}. A combination of the metrics are used to evaluate the performance of the LLMs such as BLEU, accuracy, and perplexity~\cite{zhou2023don}. It is not possible to estimate the effect of quantizing each layer of the LLM on all these metrics unless using a brute force approach which is not scalable considering the large scale of the LLMs. Therefore, we first focus on estimating the performance of LLM after quantizing each layer. To do that, we borrow the Knowledge Distillation~(KD) concept. We use the main LLM model as the teacher to supervise the quantization in the student model which is the quantized model. The important factor affecting the final performance of the model are the weights and biases of the model. Because we only quantize the weights in this paper, and we do not touch the biases, therefore, they remain unchanged. Therefore, we define the difference between weights of the teacher model and the student model as the performance metric instead of constraint~\eqref{eq:PerformanceGuarantee} as:


\begin{equation}
    \max_{w \in l} \; {|{(w^{o}_l - w^{s}_{lb})}|} \leq \delta \quad \forall \; l \in \mathcal{L}
    \label{eq:newPerformanceConstraint}
\end{equation}

\noindent where $w$ is the value of each element in feature map of layer $l$. $w^o_l$ and $w^s_{lb}$ refer to the value of each element in feature map of layer $l$ of teacher~(original) model and student model where the student one is quantized with $b$ bits, respectively. $\delta$ is the error rate. $w^s_{lb}$ is a function of quantization rate $b$ of layer $l$.
The $max$ function and the $||$ function makes the constrain \eqref{eq:newPerformanceConstraint} non-convex and the optimization problem non-linear.

\textbf{Constrain \eqref{eq:newPerformanceConstraint} Linearization}
We know that any $|N| \leq I$ is equal to $-I \leq N \leq I$ where $N$ and $I$ are just example numbers. Additionally, the $\max$ function which traverse the weights of layer $l$ can be converted to $\forall$ weights in layer $l$. 
Therefore, the constraint~\eqref{eq:newPerformanceConstraint} can be re-written as:

\begin{equation*}
    -\delta \leq {(w^{o}_l - w^{s}_{lb})} \leq \delta \quad \forall \; w \in l, l \in \mathcal{L}
\end{equation*}

And finally, we break this constraint to two constraints to incorporate the two inequalities:

\begin{equation}
    {w^{o}_l - w^{s}_{lb}} \leq \delta \quad \forall \; w \in l, l \in \mathcal{L}
\end{equation}

\begin{equation}
    {w^{o}_l - w^{s}_{lb}} \geq -\delta \quad \forall \; w \in l, l \in \mathcal{L}
\end{equation}






\textbf{Quantization Function}
Quantization can be symmetric or asymmetric. Choosing a quantization function depends on the distribution of the data in each layer. For example, for one-tailed data distributions, a symmetric function performs better while signed symmetric quantization might be suitable for distributions that are roughly symmetric about zero~\cite{wang2024model}. The data distribution of the weights affects model's performance and behavior. Therefore, knowing the distribution of weights in each layer of the LLM is of fundamental importance.




\begin{theorem}
    The optimization problem \eqref{eq:optimization} is NP-hard if and only if we have more than one LLM in the system.
    \label{theorem:np-hard}
\end{theorem}

\begin{proof}
    We prove the Theorem~\ref{theorem:np-hard} by a reduction from the NP-hard Job-shop Scheduling Without Preemption~(JSP) problem~\cite{mastrolilli2011hardness} to our problem when we have more than LLM. we are given $n$ jobs $J = {j_1, j_2, ..., j_n}$ with different processing times, which need to be scheduled on $M={m_1, m_2, ..., m_m}$ machines with varying processing power $p_m$. Each job consists of a set of operations $O_1, O_2, ..., O_n$ which need to be processed in a specific order~(known as precedence constraints).
    The objective is to minimize the makespan – the total length of the schedule~(i.e. when all the jobs have finished processing) while serving all jobs. 
    The decision variable $x_{ij}$ is defined if and only if the job $j \in J$ is assigned to machine $m \in M$. And, the decision variable $y_{oo'}$ where $o'>o$ is defined to guarantee that operation $o'$ can start if and only if operation $o$ is finished.
    We will show that a simple instance of our problem would be as hard as JSP problem. We now construct an instance of our problem. We construct $m$ edge servers with $m$ machines where each has a total communication, computation, and storage capacity of $p_m$~(i.e. $p_m = \eta_m + \gamma_m + \kappa_m$. For each job $j \in J$ construct a LLM query with operations defined as layer $l_o in \mathcal{L}$ where $l$ can be processed if and only if ${l_1, l_2, ..., l_{o-1}}$ are all processed. We assume that the compression ratio is 1 meaning that the layer $l_o$ is not compressed when assigning to machine $m$. The objective is to minimize the total length of schedule which is total computation time of processing each job on each machine plus total communication time of sending the output of the job to the next machine. We claim that the optimal solution to the constructed instance of our problem with more than one LLM gives the optimal solution to the JSP problem. This is because an algorithm that solves our problem can solve the JSP. Since JSP is NP-hard, this concludes our proof.
\end{proof}

In this paper, we focus on the case that we only have one LLM. The JSP problem with one job is not NP-hard~\cite{mastrolilli2011hardness}. Therefore, we leverage one of the off-the-shelf solvers to solve the DILEMMA problem.

%% file: Sections/Results.tex
\textbf{Environment Setup.}
We use the Python Pulp package to solve the DILEMMA problem. The DILEMMA problem is a large size problem with several parameters. Therefore, to find the effect of each important parameters, we relax other parameters to make sure that the problem is still feasible. We compare the results of LLM performance compared to the original LLM model (not quantized and centrally processed model). We use OPT-350m model~\cite{zhang2022opt} and SQuAD dataset~\cite{rajpurkar2016squad} for the evaluation. Unless otherwise stated, we set the maximum length of token equal to 128 and batch size equal to 128. We consider the minimum bit precision to 4 and set the bit steps to 2. To have a fair comparison, we divide the value of ${w^{o}_l - w^{s}_{lb}}$ by $2^b$ to have a normalized value.
To choose the best quantization method for the OPT-350m, we present the the weight distribution of OPT-350m model~\cite{zhang2022opt} in the Fig.~\ref{fig:opt350_weight_dist}. It implies that weights are mostly distributed in the range of -1 and 1. Therefore, using any symmetric/asymmetric quantization with a round function only increases error. We use truncation method for the quantization technique. 

\begin{figure}
    \centering
    \includegraphics[width=0.9\linewidth]{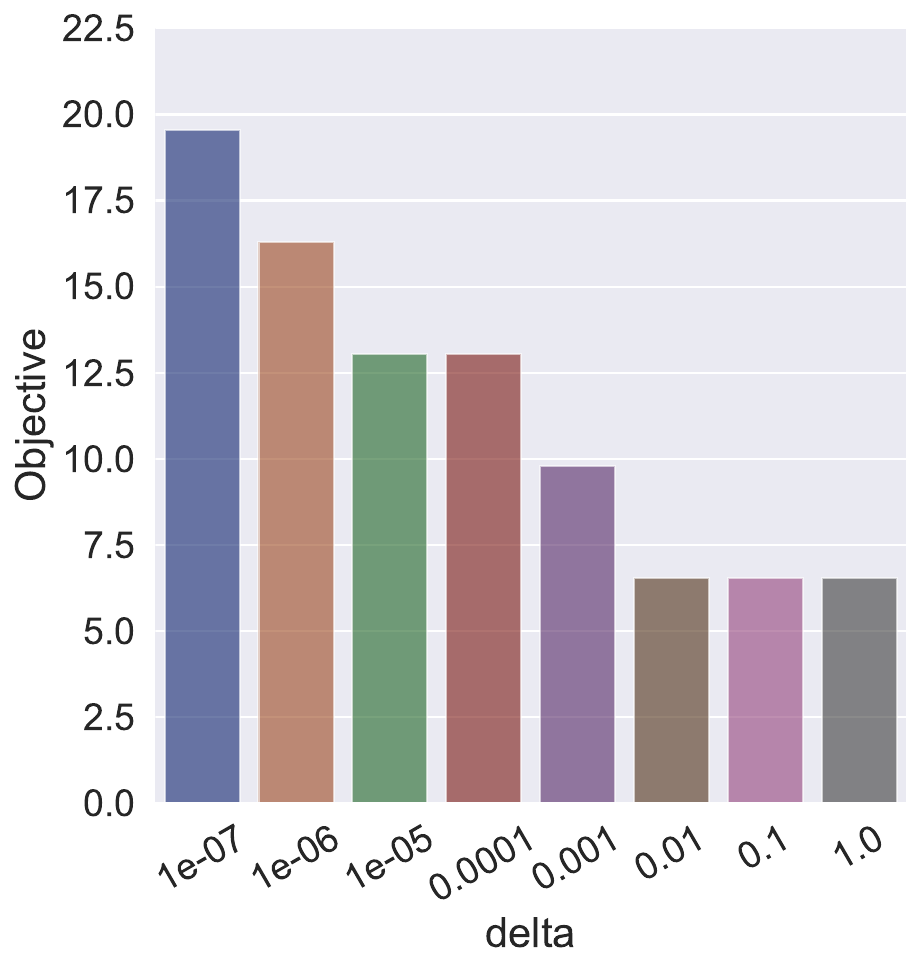}
    \caption{Objective value based on different $\delta$ values}
    \label{fig:obj_delta}
\end{figure}

\begin{table}[]
    \footnotesize
    \centering
    \caption{The value of loss and perplexity of the OPT-350m model using DILEMMA scheme based on different $\delta$ values. The loss value of the original model is 0.0591, its BLEU is 0.2951, and the its perplexity is 1.0609. 
    }
    \label{tab:delta}
    \begin{tabular}{rcccc}
        \toprule
        & \multicolumn{3}{c}{LLM Performance Metric} \\
        \cmidrule{2-4}
        $\delta$ Values & Loss & Perplexity & BLEU & Quant. Ratio\\
        \midrule
        1e-7 &  0.0591 & 1.0609 & 0.2949 & 37.50\%\\
        1e-6 & 0.0591 & 1.0609  & 0.2945 & 31.25\%\\
        1e-5 &  0.0591  & 1.0609 & 0.2942 & 25.00\%\\
        0.0001 & 0.0591 & 1.0609 & 0.2942 & 25.00\%\\
        0.001 &  0.0591 & 1.0609 &  0.3050 & 18.75\%\\
        0.01 & 0.0605 & 1.0623 & 0.2939 & 12.50\%\\
        0.1 & 0.0605 & 1.0623 & 0.2939 & 12.50\%\\
        1.0 & 0.0605 & 1.0623 & 0.2939 & 12.50\%\\
\bottomrule
    \end{tabular}
    \label{table:delta_performance}
\end{table}

\textbf{Effect of $\delta$.}
We relax the storage constraint~\eqref{eq:constraintstorage} to find the effect of $\delta$ on the DILEMMA problem.
Fig.~\ref{fig:obj_delta} represents the objective value of the DILEMMA problem for different $\delta$ value. As the $\delta$ decreases, the objective increases which means an increase in the total completion time of the inference task. Although the completion time increases, the performance of the quantized model is better compared to the case where $\delta$ is equal to 1 as presented in the Table.~\ref{table:delta_performance}. The Table.~\ref{table:delta_performance} implies that a good level of the LLM performance can be achieved even with not a very strict delta value such as $1e-5$. Additionally, when the $delta$ decrease, the performance increase. 
We define the quantization ratio as the $\frac{\sum_l{p_l \; b_l}}{\sum_l{p_l \; pr_l}} \times 100$ where $p_l$ is the total number of weights of layer $l$, $b_l$ is the precision of the weights of layer $l$ after quantization, and the $pr_l$ is the precision of layer $l$ weights in the original form (i.e. before quantization). The results states that the higher the $\delta$ value is, the more the quantization rate is, and consequently, less total completion time but with a low performance degradation.

\begin{figure}
    \centering
    \includegraphics[width=0.9\linewidth]{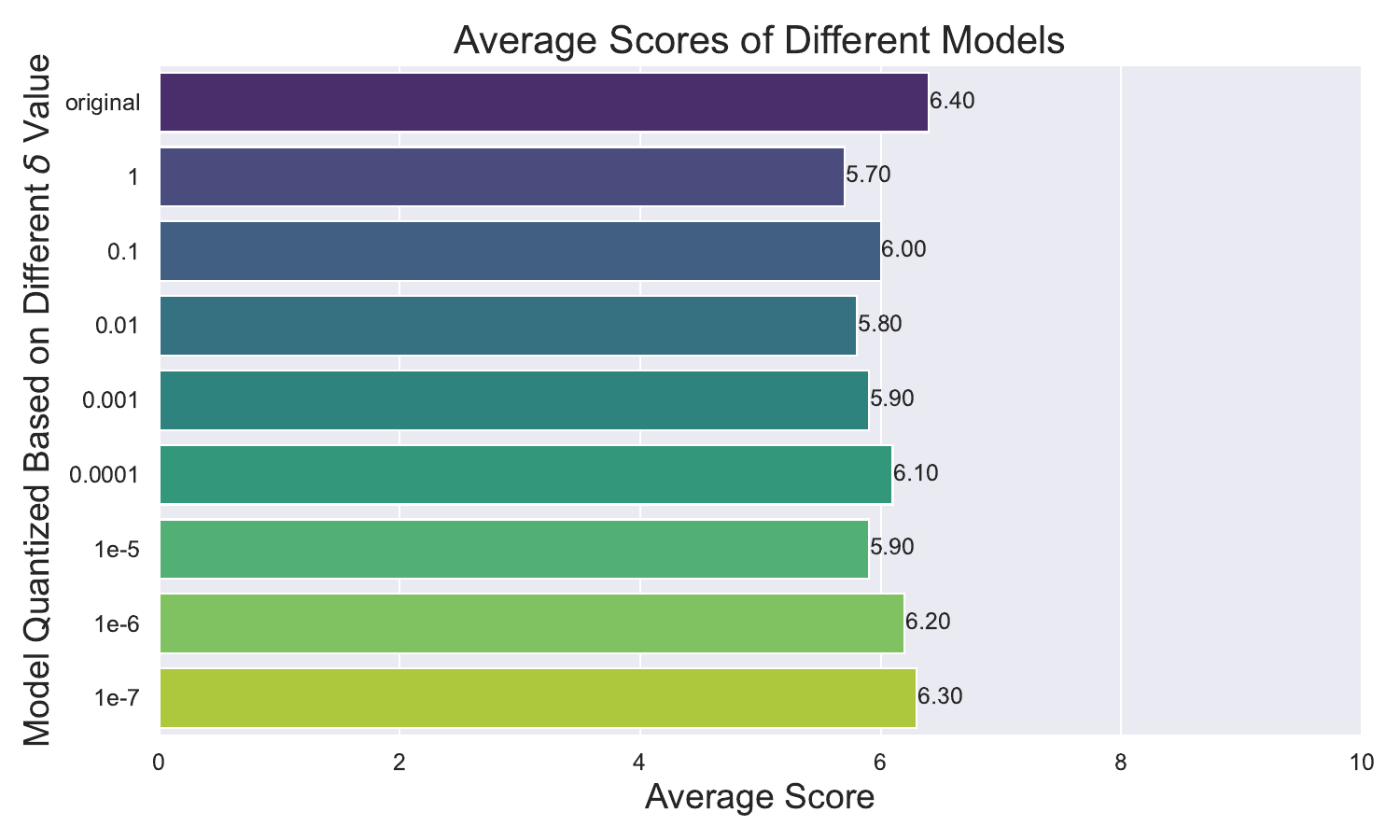}
    \caption{Models scored by GPT-4 based on their answers to SQuAD questions dataset}
    \label{fig:model_scores_barplot}
\end{figure}

Next, we randomly chose 1000 questions from SQuAD dataset~\cite{rajpurkar2016squad} and used all the quantized OPT350m models represnted in Table.1 along with the original model to answer these questions. Then, we asked GPT-4 to score each questions and answers by each model based how relevant, correct, clearness, and fluent they are. The range of the scores are from 0 to ten where zero is the worst and ten is the best. Fig.~\ref{fig:model_scores_barplot} presents the scores provided by GPT-4. It implies that quantizing the model with strict $\delta$ value provides models with more similar answers compared to the original LLM.

\begin{figure}
    \centering
    \includegraphics[width=0.9\linewidth]{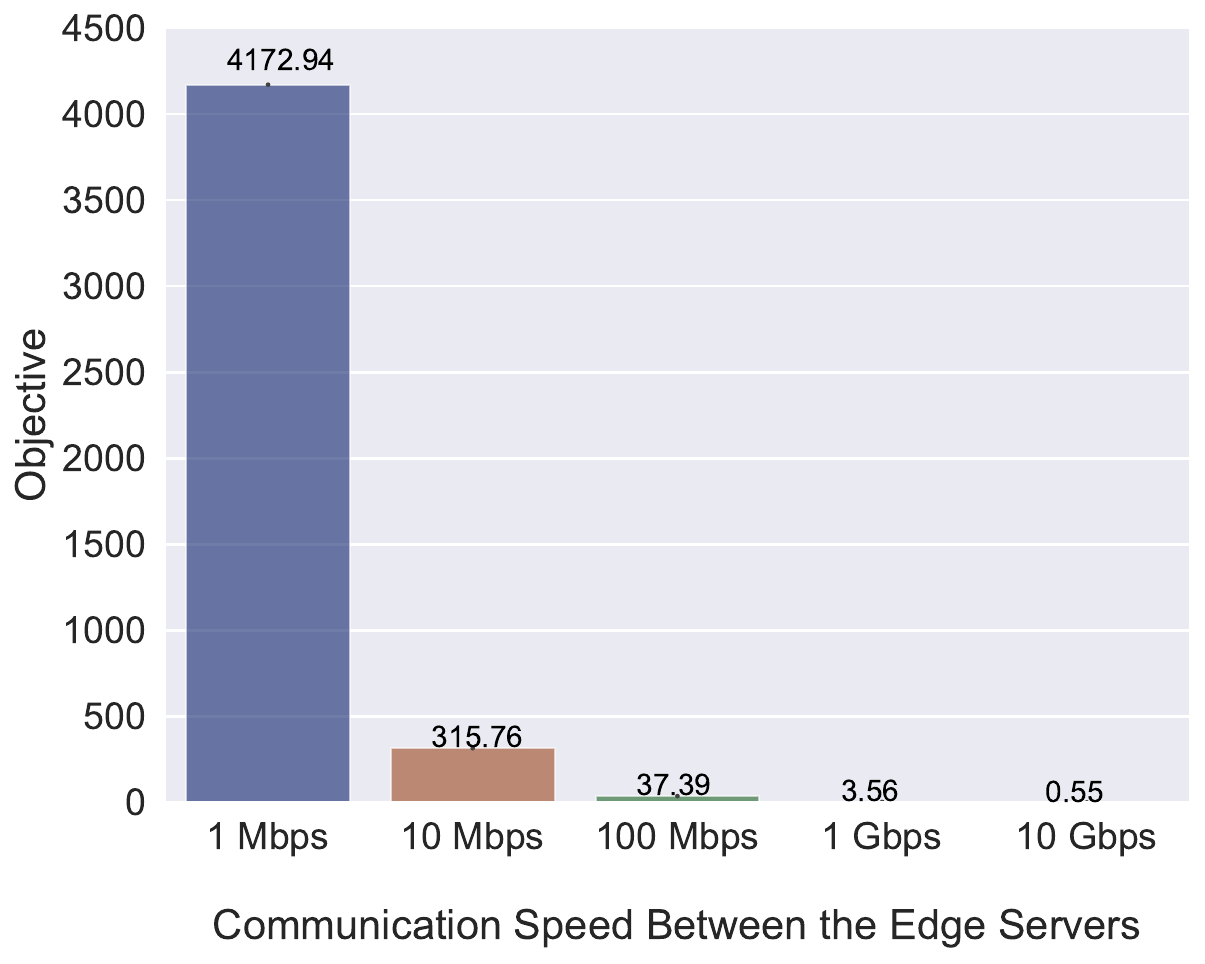}
    \caption{The impact of $\Omega_{ij}$ on the objective value}
    \label{fig:obj_com_barplot}
\end{figure}

\textbf{Effect of Communication Speed Between the Edge Servers on the Objective Value.}
We run the next tests for 100 Monte Carlo runs. We generate the communication speed between the edge servers using the random number between from the range of $[\text{min}_{\Omega_{ij}}, 10 * \text{min}_{\Omega_{ij}}]$.
We change the  $\text{min}_{\Omega_{ij}}$, from 1 Mbps to 10 Gbps to find the impact of communication speed between the edge layers on the objective value of the DILEMMA problem. The results are plotted in the Fig.~\ref{fig:obj_com_barplot} implying that with increasing the communication speed between the edge servers, the objective decreases, meaning an improvement in the total completion time of the autoregressive inference task.

\begin{figure}
    \centering
    \includegraphics[width=\linewidth]{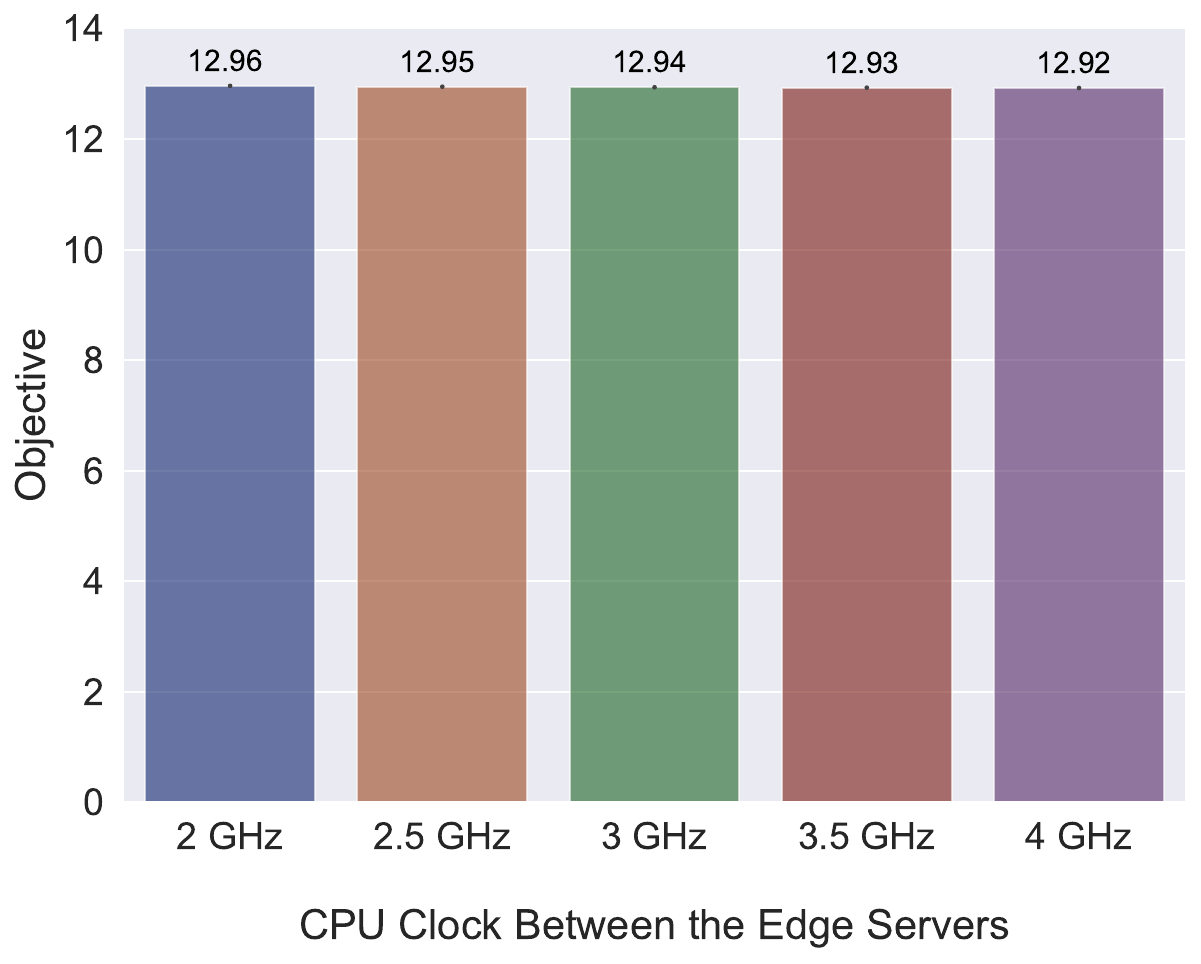}
    \caption{The impact of $CCS_{i}$ on the objective value}
    \label{fig:obj_cpu_barplot}
\end{figure}

\textbf{Effect of Computation Capacity of the Edge Servers on the Objective Value.}
Next, we investigate the effect of $\text{CCS}_i$, the CPU clock of each edge server $i \in \mathcal{M}$, on the objective value. Fig~.\ref{fig:obj_cpu_barplot} represents the impact of $\text{CCS}_i$ on the objective value implying the objective value decreases when the $\text{CCS}$ increases; i.e. improving in completion time.

%% file: Sections/Conclusion.tex
This paper presents DILEMMA, a framework for the joint optimization of LLM layer placement and quantization in edge computing systems. By leveraging layer-wise quantization and knowledge distillation, the framework addresses the challenges posed by limited computational, communication, and storage resources in edge servers. Experimental results validate the efficiency of the proposed approach, showing that it achieves a balance between LLM performance and resource utilization while reducing model size by $87.5\%$. In future work, we will explore extending this approach to multiple LLMs and dynamic edge network conditions while considering multimodal LLMs.